\documentclass[letterpaper, 10 pt, conference]{ieeeconf}  

\IEEEoverridecommandlockouts                              

\usepackage{soul}
\usepackage{booktabs}
\usepackage{todonotes}
\usepackage{algorithm}
\usepackage{algpseudocode}
\usepackage{amsmath}
\usepackage{relsize}
\usepackage{subfig}
\usepackage{hyperref}
\usepackage{multirow}
\usepackage{graphicx}

\title{\LARGE \bf{VeREFINE: Integrating Object Pose Verification\\with Physics-guided Iterative Refinement}}

\author{Dominik Bauer, Timothy Patten and Markus Vincze
\thanks{This work is partially funded by the TU Wien Doctoral College TrustRobots and the Austrian Science Foundation (FWF) under grant agreement No. I3968-N30 HEAP and No. I3969-N30 InDex.}%
\thanks{The authors are with the Automation and Control Institute, TU Wien, 1040 Vienna, Austria {\tt\small\{bauer,patten,vincze\} @acin.tuwien.ac.at}.}%
}

\begin{document}

\maketitle
\thispagestyle{empty}
\pagestyle{empty}


\begin{abstract}
Accurate and robust object pose estimation for robotics applications requires verification and refinement steps. %
In this work, we propose to integrate hypotheses verification with object pose refinement guided by physics simulation. This allows the physical plausibility of individual object pose estimates and the stability of the estimated scene to be considered in a unified optimization. The proposed method is able to adapt to scenes of multiple objects and efficiently focuses on refining the most promising object poses in multi-hypotheses scenarios. %
We call this integrated approach \emph{VeREFINE} and evaluate it on three datasets with varying scene complexity. The generality of the approach is shown by using three state-of-the-art pose estimators and three baseline refiners. Results show improvements over all baselines and on all datasets.
Furthermore, our approach is applied in real-world grasping experiments and outperforms competing methods in terms of grasp success rate.
Code is publicly available at \href{https://www.github.com/dornik/verefine}{github.com/dornik/verefine}.
\end{abstract}

%
\IEEEpeerreviewmaketitle


\section{Introduction}
Autonomous robots need to interact with their physical environment to fulfill a plethora of tasks. This requires the manipulation of individual objects, for example, to fetch an item or to stow it away. A popular approach to enable such manipulations uses object pose estimation and grasp pose annotation~\cite{Tremblay2018_DeepOP, Srinivasa2010, Chitta2012, Wang2019}.
Previous work on object detection and pose estimation achieves high accuracy on popular datasets such as \texttt{LINEMOD}~\cite{Hinterstoisser2012} or \texttt{YCB-VIDEO}~\cite{Xiang2017}. However, the performance of these algorithms deteriorates when the objects' 3D models are inaccurate or lighting and viewing conditions change~\cite{Loghmani2018,Ammirato2017}.
To deal with this problem, hypotheses verification and object pose refinement are commonly used in object pose estimation pipelines.

\begin{figure}[!t]
\centering
\setlength{\tabcolsep}{2pt} 
\begin{tabular}{cc}
\raisebox{-0.69\height}[0pt][0pt]{\includegraphics[trim=0 0 0 50,clip,width=0.48\linewidth]{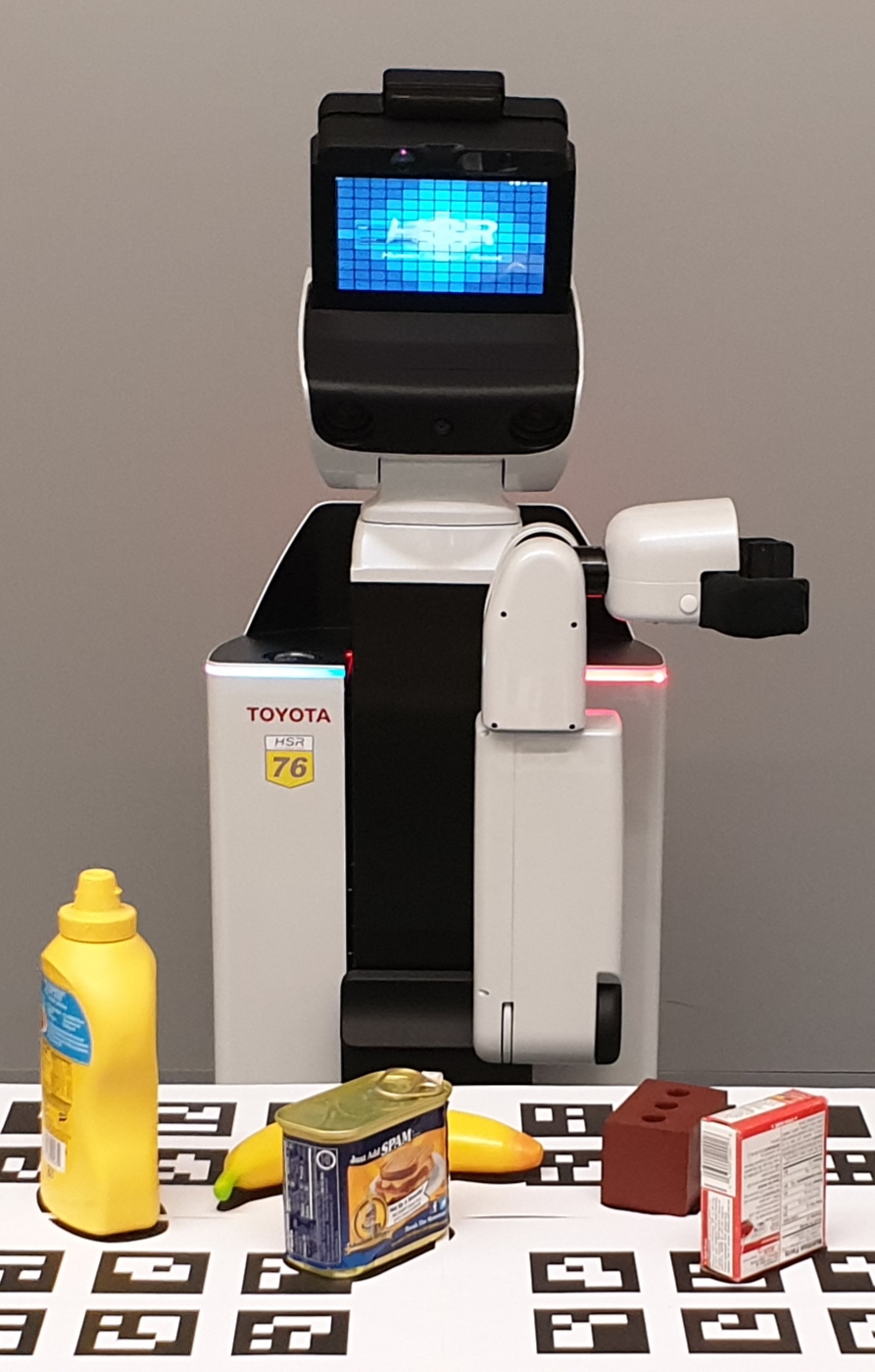}} & \includegraphics[trim=100 220 120 140, clip,width=0.48\linewidth]{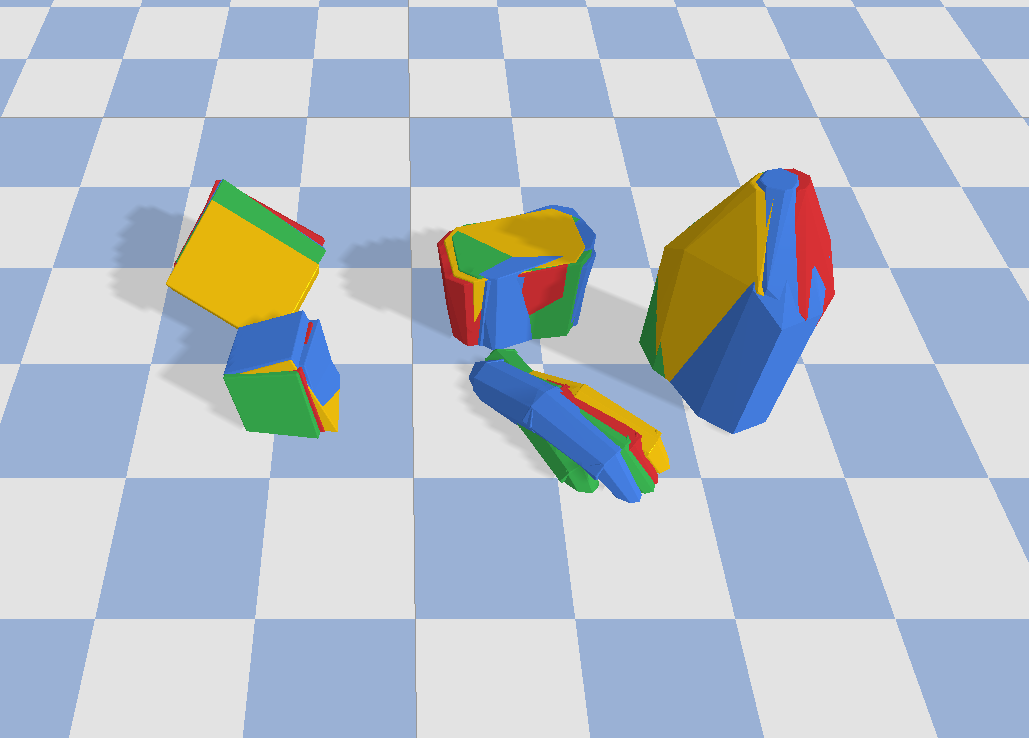} \\
 & \includegraphics[trim=80 190 110 120, clip,width=0.48\linewidth]{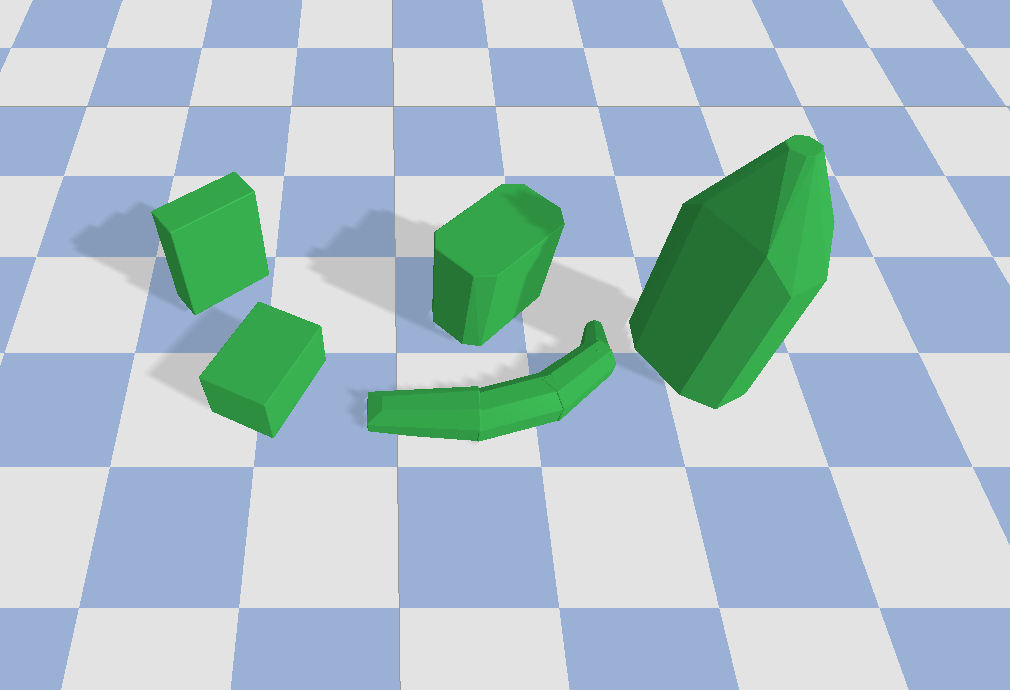} \\
 & \includegraphics[trim=160 155 110 95, clip, width=0.48\linewidth]{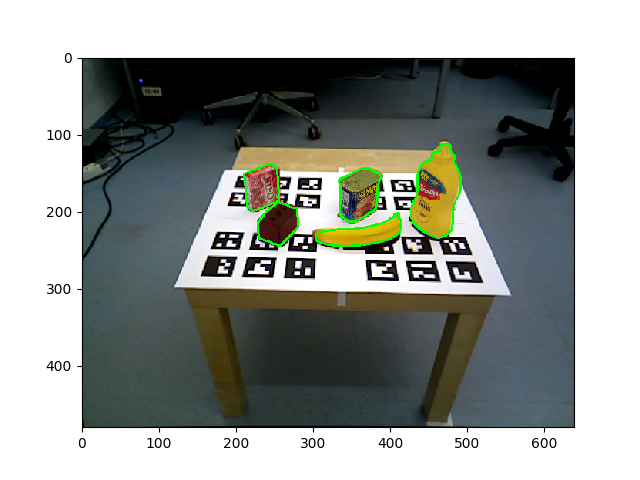} \\
\end{tabular}
\setlength{\tabcolsep}{6pt} 
\caption{Grasping \texttt{YCB-VIDEO} objects with a Toyota HSR. Initial pose estimates in simulation environment (top) are improved using VeREFINE (mid and bottom).}
\label{fig:teaser}
\end{figure}

The idea of hypotheses verification is to evaluate the fit of different estimates with the observed scene: The best fitting estimates are selected and estimates below a threshold are pruned. While this improves accuracy and reliability it also introduces the problem of increased complexity arising from the number of possible combinations in multi-object scenes.
The usability of such approaches in robotics is limited by their runtime. For example, Mitash et al.~\cite{Mitash2018} combine object pose verification with physics simulation, resulting in frame times of up to 30s. Krull et al.~\cite{Krull2017} integrate object pose refinement and verification with reinforcement learning to efficiently allocate a refinement budget. However, the authors report a frame time of up to 34s.
 
In contrast to hypothesis verification that only accepts or rejects object pose estimates, object pose refinement improves the estimates themselves. This is achieved by minimizing the discrepancy between the observed scene and the object in an estimated pose. The most popular pose refinement method is the Iterative Closest Point (ICP) algorithm~\cite{Besl1992}.
However, if the initial estimate or the visual observation are inaccurate, ICP converges to a local minimum (wrong pose) or even diverges.
Alternatively, physics simulation has been used to ensure plausibility and improve accuracy \cite{Mitash2018,Furrer2017} of object pose estimates. But applying physics simulation to objects is an unstable process. It may cause objects to topple over and create worse estimates given the inaccuracy of the simulated environment.

The goal of both verification and refinement is to maximize the fit of the estimate to the observed scene.
We hypothesize that, by integrating these approaches into one step, we are able to improve the overall accuracy of the pose estimates, while achieving more graceful degradation by limiting divergence of individual strategies.
To this end we present VeREFINE, an integrated approach that combines hypotheses \textit{Ver}ification, object pose \textit{Refine}ment and physics simulation into a unified framework. Our contributions are

\begin{itemize}
    \item improving accuracy by integrating refinement with physics simulation in an iterative loop,
    \item improving robustness by efficient rendering-based verification of object pose estimates,
    \item improving accuracy and runtime using regret minimization to exploit promising hypotheses, and
    \item the combination into reliable scene-level refinement and verification for multi-object scenes.
\end{itemize}

We evaluate our framework on three publicly available datasets, \texttt{Extended APC} \cite{Mitash2018}, \texttt{LINEMOD} \cite{Hinterstoisser2012} and \texttt{YCB-VIDEO}~\cite{Xiang2017}, and out-perform state of the art in pose estimation and refinement in terms of robustness and accuracy.
We compare to the related approach by Mitash et al.~\cite{Mitash2018}, achieving a significant reduction in runtime while increasing the accuracy of the pose estimates.
We demonstrate the robustness of our method with respect to initial pose errors and missing depth values due to occlusion and material properties.
Finally, we evaluate the proposed framework in a robotic grasping experiment resulting in significantly increased success rates compared to other methods.

After reviewing related work in Sec.~\ref{sec:related_work}, we discuss the refinement methods in Sec.~\ref{sec:single} and the complete VeREFINE approach in Sec.~\ref{sec:vf}. Sec.~\ref{sec:experiments} presents experiments and results. Sec.~\ref{sec:conclusion} concludes the paper.


\section{Related Work}\label{sec:related_work}
The proposed approach builds on previous work in hypotheses verification, object pose refinement and their combination with physics simulation.

Hypotheses verification approaches for object pose estimation show that considering multiple pose hypotheses per object improves overall estimation performance. Drost et al.~\cite{Drost2010} use a clustering-based verification stage to refine pose estimates. In~\cite{Vidal2018}, a pool of 200 object pose hypotheses is generated using a Point Pair Features (PPF) pipeline. Each hypothesis is refined using Projective ICP and a two-step verification to determine the best estimate.
In~\cite{Xiang2017}, an initial estimate is perturbed to generate a set of hypotheses for better coverage of the solution space. All hypotheses are refined before scoring and selection. In contrast, Wang et al.~\cite{Wang2019} estimate a pose confidence score jointly with per-pixel object pose estimates. The highest scoring estimate is selected and refined. 
Krull et al.~\cite{Krull2017} train a CNN to predict two different hypotheses scores for use during refinement and for the selection of the final estimate.
On the scene level, a scoring function that considers geometrical cues, clutter and conflicting hypotheses for multiple objects is proposed in~\cite{Aldoma2016}.
For efficient evaluation of the search space, \cite{Bauer2019icvs} consider equivalent combinations of hypotheses to reduce the search tree to a directed acyclic graph and explore using Monte Carlo Tree Search (MCTS). Physics simulation is incorporated in MCTS to additionally consider the supporting relations between objects in~\cite{Mitash2018,Bauer2019oagm}. We propose to apply rendering-based verification to guide refinement, allowing refinement steps to be allocated to promising hypotheses. This naturally extends to multi-object scenes, which reduces the solution space as compared to search-based methods.

Previous work on object pose refinement exploit depth, RGB and object segmentation as input modalities. A seminal approach is ICP \cite{Besl1992}. 
More recently, deep learning approaches for object pose refinement have been proposed. RGB-based methods render intermediary object pose estimates and use CNNs to compute a pose update~\cite{Li2018,Manhardt2018,Zakharov2019}.
The refinement method by Wang et al.~\cite{Wang2019} requires RGB-D images and instance segmentation as input. The depth cues are processed using PointNet~\cite{Qi2017} and combined with the RGB-based features from a CNN.
We show that our proposed approach is applicable to both learning-free and learning-based methods. It boosts their performance by improving initial estimates using physics simulation and guides refinement through rendering-based verification.

Application of physics reasoning and simulation in related vision tasks indicates that it creates strong cues for object pose and admissible scene configurations.
The segmentation method by Jia et al.~\cite{Jia2014} uses rule-based physical stability reasoning to combine or split candidate patches, represented by bounding boxes, to generate physically plausible scenes. A similar reasoning is applied to voxelized scene representations to segment and estimate the shape of objects in~\cite{Zheng2015}.
In a robotics context, Furrer et al.~\cite{Furrer2017} show the benefit of using physics simulation for object stacking. They propose a method for determining the target pose of irregular stones such that a structurally stable stack can be built by a robot. 
Mitash et al.~\cite{Mitash2018} use physics-based verification and MCTS for object pose estimation given multiple hypotheses in multi-object scenes. 
For each hypotheses combination, this approach runs one iteration of Trimmed ICP and a physics simulation, making it sensitive to the estimation of the supporting plane and the physical properties of the simulated objects.
Our proposed solution of interleaving physics simulation and refinement is more robust to these challenges and prevents diverging simulation. We allow more promising estimates to be refined multiple times while saving these additional iterations on less promising estimates.
Moreover, in~\cite{Mitash2018}, a solution is processed one object after another. Feedback on the impact on the overall solution quality is given by a scene-level reward but only allows to select among the refined hypotheses. In contrast, by incorporating the scene-level feedback in the refinement process, our approach adapts the estimates to the overall solution.
Furthermore, the approach of~\cite{Mitash2018} needs to grow a search tree of combinations of hypotheses, spending expensive refinements on exploring the search space. More efficiently, our approach uses an object-based representation of the search space, which is initialized using a rendering-based verification score. Thereby, no additional computation needs to be spent on initial exploration of the search space.


\section{Integrating Hypotheses Verification with Physics-guided Iterative Refinement}\label{sec:single}

The goal of this work is to accurately and robustly explain scenes of varying complexity in terms of object poses for applications such as robotic grasping. An RGB-D observation, instance detection, instance segmentation and a set of initial object pose estimates are assumed to be given.

\begin{figure}[!t]
\centering
\vspace{1ex}
\includegraphics[width=0.8\linewidth]{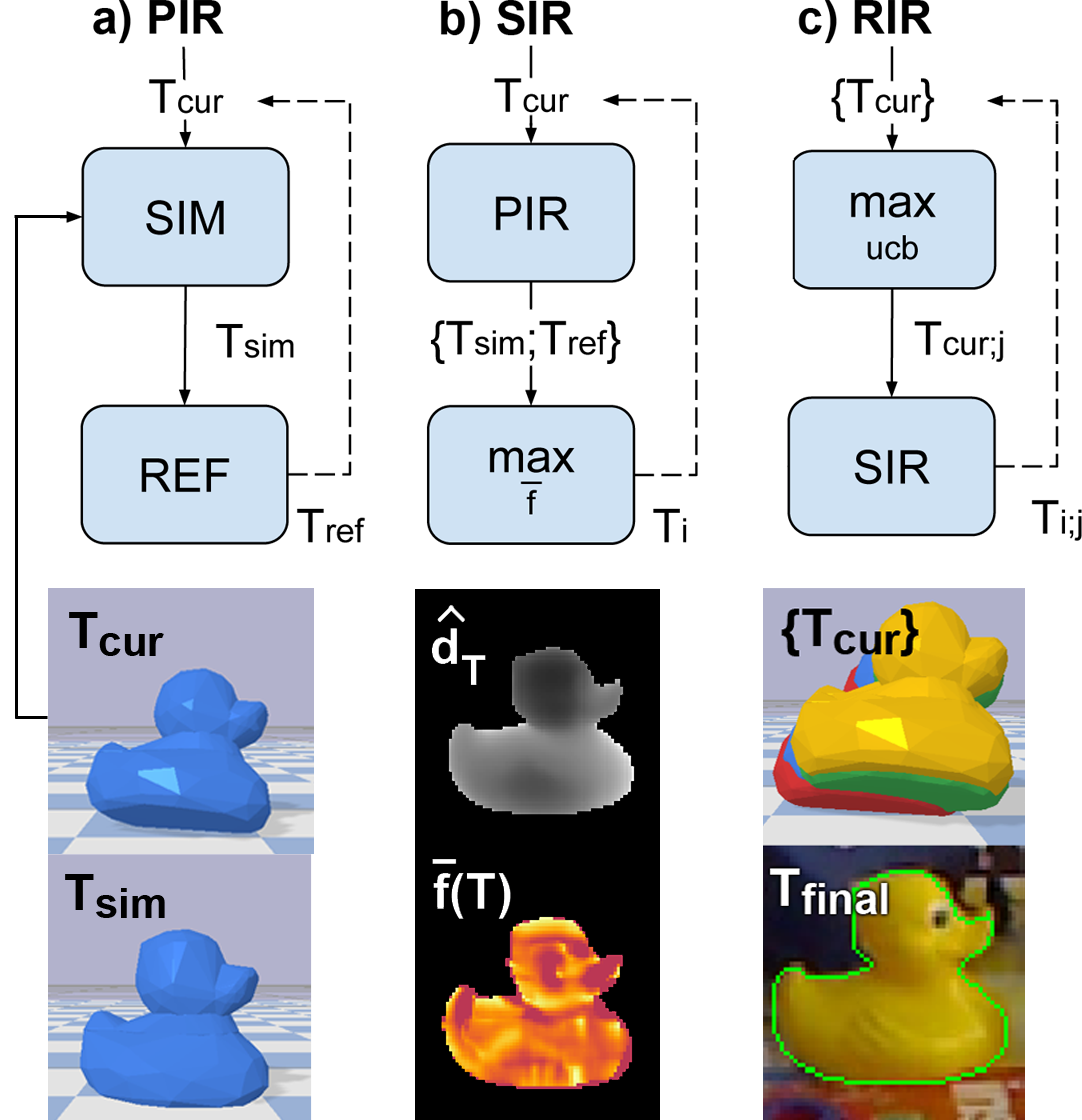}
\caption{Proposed integration approaches given a simulation environment and an initial object pose estimate ($T_{cur}$). (a) Integration of physics simulation (SIM) and iterative refinement (REF) into Physics-guided Iterative Refinement (PIR). (b) Supervision using verification score $\Bar{f}$ (SIR). (c) Regret minimization using UCB score (RIR).}
\label{fig:schematic_single-object}
\end{figure}

In the following, we present the building blocks of our VeREFINE approach by considering increasingly complex scenarios. For individual objects, we propose an iterative physics-guided refinement loop (Sec.~\ref{sec:pir}). To improve the robustness of this approach, supervision of the refinement loop through rendering-based verification is presented (Sec.~\ref{sec:sir}). Given multiple estimates, a regret minimization approach is introduced to efficiently allocate refinement towards promising estimates (Sec.~\ref{sec:rir}). We extend the discussed methods to consider multi-object scenes with multiple initial estimates each, where occlusion and support relationships between objects need to be considered (Sec.~\ref{sec:vf}).

\subsection{Physics-guided Iterative Refinement (PIR)}\label{sec:pir}
Object pose refinement methods depend heavily on the quality of the initial estimates.
In contrast to previous approaches that apply physics simulation as a post-hoc step after iterative refinement \cite{Mitash2018}, we propose to interleave object pose refinement and physics simulation in a Physics-guided Iterative Refinement (PIR) loop, illustrated in Figure \ref{fig:schematic_single-object}a.
The physical plausibility of the initial estimate used for refinement is improved using simulation, helping the refinement to relate the correct parts of the model to the observation. The iterative feedback loop allows the refinement to, in turn, initialize physics simulation with more accurate estimates, thus limiting divergence.

In each iteration, the current object pose estimate $T_{cur}=[R_{cur},t_{cur}]$ initializes the object in the simulation environment, shown in Figure \ref{fig:schematic_single-object}a (mid). In the simplest case, the environment consists of a supporting plane. In more complex scenes, it also includes other estimated objects. The simulation is progressed and the resulting object pose $T_{sim}$ is returned. As indicated in Figure \ref{fig:schematic_single-object}a, only the orientation part $R_{sim}$ is used to update the estimate. This is motivated by the observation that, when physics simulation leads to large displacements, it causes the iterative refinement to lose track of corresponding object parts.
We found only using the orientation contains this divergent behavior while still improving the refinement process. The estimate $[R_{sim}, t_{cur}]$ is used to initialize an iteration of the object pose refinement algorithm that returns the final estimate $T_{ref}$ after one iteration of PIR.


\subsection{Supervised Iterative Refinement (SIR)}\label{sec:sir}
Due to divergent behavior in physics simulation or iterative refinement, the final estimate after applying these methods might generate a worse explanation of the observation than the initial or intermediary estimates. We solve this by continuously evaluating the observation fit of the intermediary estimates. This integration of verification into the refinement process allows us to supervise divergent behavior and select the best fitting estimate as the final one. The verification score $\Bar{f}$ measures the observation fit and is computed from the average discrepancy between the estimate and the observation in terms of depth and surface normals, given by
\begin{equation}\label{eq:score}
\begin{split}
    \Bar{f}(T) &= \frac{1}{2} \left( \frac{1}{N} \mathlarger{\sum\limits^{N}} f_d(T) + \frac{1}{N} \mathlarger{\sum\limits^{N}} f_\mathbf{n}(T) \right) \\
    f_d(T) &= 
    \begin{cases}
        1 - \frac{|d - \hat{d}_T|}{\tau},& \text{where $|d - \hat{d}_T| < \tau$} \\
        0,& \text{otherwise}
    \end{cases} \\
    f_\mathbf{n}(T) &= 
    \begin{cases}
        1 - \frac{1-{\mathbf{n} \cdot \mathbf{\hat{n}}_T}}{\alpha},& \text{where $1-{\mathbf{n} \cdot \mathbf{\hat{n}}_T} < \alpha$} \\
        0,& \text{otherwise}
    \end{cases}
\end{split}
\end{equation}
where $d$ is a valid depth value and $\mathbf{n}$ is a corresponding surface normal in the segmented scene. The $N$ corresponding values of the rendered estimate are $\hat{d}_T$ and $\hat{\mathbf{n}}_T$. Parameters $\tau$ and $\alpha$ are soft thresholds for the maximal admissible discrepancy. Figure \ref{fig:schematic_single-object}b (bottom) shows an example of $\Bar{f}$ applied to an estimate.

In each PIR iteration $i$, we evaluate the estimates returned by physics simulation $T_{i;sim}$ and refinement $T_{i;ref}$ and proceed with the estimate that achieves the better score. After the last iteration, the final estimate $T$ that gives the best score $\Bar{f}$ overall is selected from all processed estimates. As such, Supervised Iterative Refinement (SIR) covers cases where the individual approaches could diverge.

The supervision requires evaluations of $\Bar{f}$ for $T_{i;sim}$ and $T_{i;ref}$ each iteration. To enable fast evaluation, computations are carried out on the GPU in two rendering passes using OpenGL. The first pass writes $\hat{d}_T$ and $\hat{\mathbf{n}}_T$ to a texture. The second pass uses this texture and the observation to compute $f_d$ and $f_n$. The summed values of $N$, $f_d$ and $f_n$ are read-back from a higher-level mipmap, drastically reducing the read-back time. The final averaging is done on the CPU and yields $\Bar{f}$. In our experiments, one evaluation of $\Bar{f}$ using a NVIDIA GTX 1080Ti takes 1-2ms. This is a significant speed-up compared to 7-9ms when reading-back the full depth and normal information from the GPU to evaluate $\Bar{f}$ on the CPU.


\subsection{Regret-minimizing Iterative Refinement (RIR)}\label{sec:rir} 
Considering multiple pose hypotheses per object raises the questions: On which hypotheses to spend refinement steps and which hypothesis to select in the end. Promising hypotheses should be exploited by applying more refinement steps while other hypotheses should still be explored to find better candidates. 

We propose a Multi-armed Bandit (MAB) to model this exploitation-exploration problem, where the pull of arm $j$ represents running one SIR iteration for hypothesis $j$. The Upper Confidence Bound policy (UCB) \cite{Auer2002} minimizes the regret of choosing a sub-optimal arm of a MAB with respect to a given reward. In each iteration, the arm with maximal $ucb_j$ is selected according to
\begin{equation}
    ucb_j = \mu_j + c \cdot \sqrt{\frac{\ln{p}}{n_j}}
\end{equation}
where $\mu_j$ is the mean reward of playing arm $j$, $p$ is the total number of plays and $n_j$ is the number of times the arm has been played. $c$ is a parameter of the algorithm that controls the balance of exploitation and exploration.

In our approach, illustrated in Figure \ref{fig:schematic_single-object}c, the verification score $\Bar{f}$ is chosen as reward function. Applying the UCB policy to the resulting reward statistics efficiently allocates a fixed refinement budget, spending more refinements on promising hypotheses while saving refinements on those that have a low $\Bar{f}$. This results in the same total amount of refinements but in a regret-minimizing way.
The resulting Regret-minimizing Iterative Refinement (RIR) procedure starts by ranking the initial estimates based on $\Bar{f}$. For each subsequent RIR iteration, SIR is applied and the verification score is used as reward signal. As with SIR, the final selection is based on the observation fit across all encountered estimates.

The formulation based on a MAB and a rendering-based score allows our approach to be quickly applied to new datasets and can be used to extend existing and future refinement methods. In contrast, the related approach in~\cite{Krull2017} uses reinforcement learning and a CNN-based verification score regression, which need to be expensively re-trained.


\section{Physics Simulation and Regret Minimization in Cluttered Multi-Object Scenes}\label{sec:vf}
In cluttered multi-object scenes, the proposed verification score and physics simulation need to deal with occlusions and support relationships. Thus, the order in which objects are considered is important. Moreover, with each of the $N$ objects having $n$ hypotheses, the number of combinations of hypotheses grows exponentially. To tame this problem, we discuss clustering strategies to reduce the number of combinations that need to be considered and present two approaches to efficiently evaluate the remaining search space.

\begin{figure}[!t]
\vspace{1.0ex}
\centering
\includegraphics[width=\linewidth]{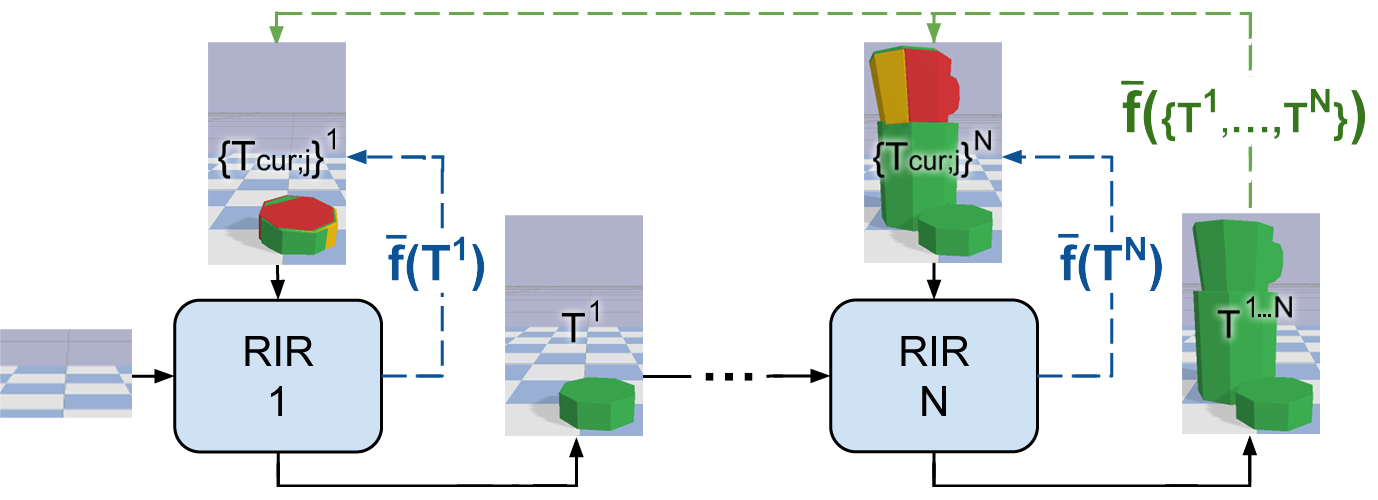}
\caption{Proposed approaches for cluttered multi-object scenes.
The best estimate per object is added to the simulation environment used for the subsequent objects, allowing consideration of occlusions and support relationships. VF$_b$ (blue) fully refines each object using the object fit as reward. VF$_d$ (green) repeats this process iteratively, refining each object only once per iteration and uses the scene fit as reward.}
\label{fig:schematic_multi-object}
\end{figure}

\subsection{Object Clustering and Dependency Graph}
Mitash et al. \cite{Mitash2018} isolate objects that might interact based on the segmented point clouds. This reduces the number of objects that need to be jointly considered and thus the number of combinations. Furthermore, they argue that not all combinations of objects have to be considered. Instead, occlusion and support relationships between objects are used to compute a dependency list. A search tree is built from this list, where at layer $i$, object $i$ is represented by all of its $n$ hypotheses. This yields a tree of $(n^{N+1}-1)/(n-1)$ nodes. For a scene of 5 objects with 5 hypotheses each this produces a search tree of 3905 nodes (excluding the root node).

In contrast, we address more general scenarios by explicitly considering ambiguous dependencies, for example, the case where an object is occluded by another object but also supporting the same object. To resolve such ambiguities, we first decompose the independent clusters into support dependency lists. The first object in each support dependency list, the base object, is in contact with the ground plane and supports the remaining objects in the list. The support dependency lists are then ordered front-to-back based on their respective base objects.
Instead of using the resulting dependency list to grow a search tree using MCTS as in~\cite{Mitash2018}, we exploit our single-object approaches to reduce the solution space and allow for iterative refinement on a scene level. The proposed representation requires only $N \cdot n$ nodes to represent the same search space as before -- or only 25 instead of 3905 nodes in the example.

\subsection{VeREFINE breadth (VF$_b$)}\label{sec:vfb}
Our first approach is to explore all object hypotheses by representing each object in the dependency list and its hypotheses using a RIR bandit. The scene is incrementally built by computing the best estimate for the considered object in the current environment. The object is added to the environment with the computed pose, allowing more accurate estimation of the next objects' poses.
We call this approach of first exploring all hypotheses per object \textit{VeREFINE breadth} (VF$_b$), shown in Figure \ref{fig:schematic_multi-object}~(blue). This results in $N$ RIR bandits with $n$ nodes each. 

\subsection{VeREFINE depth (VF$_d$)}\label{sec:vfd}
An alternative approach, and to introduce a feedback loop that is missing in VF$_b$, is to iterate through the dependency list. For each iteration, the objects' RIR bandits are progressed only once. The best known hypotheses after each iteration are evaluated as a complete scene. The resulting scene fit is computed by $\Bar{f}$ and is used as reward for selected hypotheses instead of the per-object reward. Thereby, hypotheses that contribute to a better overall scene fit are selected more often. This scene-first approach, called \textit{VeREFINE depth} (VF$_d$), is illustrated in Figure~\ref{fig:schematic_multi-object}~(green).
As the procedure results in a changing reward distribution, the UCB policy is replaced with Discounted-UCB (D-UCB)~\cite{Kocsis2006}. The reward and plays statistics are discounted by a small factor each iteration, which reduces the impact of previous iterations and adapts to a changing reward distribution over time. This is shown to reduce the cumulative regret of the D-UCB policy as compared to UCB for abruptly and continuously changing reward distributions~\cite{Garivier2011}.

The RIR bandits are initialized using the rendering-based verification score as in the single-object scenario, acting as a heuristic in the first iteration through the dependency list to select better initial estimates.
Therefore, instead of spending refinement steps to grow the search tree as in the MCTS-based approach~\cite{Mitash2018}, both our proposed approaches immediately and efficiently allocate refinement steps to more promising estimates.


\section{Experiments}\label{sec:experiments}

This section presents the evaluation of VeREFINE on the \texttt{Extended APC} (xAPC), \texttt{LINEMOD} (LM) and \texttt{YCB-VIDEO} (YCBV) datasets. Improvement over state-of-the-art refinement methods is shown by comparison with Iterative Closest Point (ICP) and DenseFusion Refinement (DF-R). For pose estimation, we use Point Pair Features (PPF) and  DenseFusion (DF). In addition, we compare against the approach presented by Mitash et al.~\cite{Mitash2018} (PHYSIM-MCTS). It uses Super4PCS (PCS) and hypotheses clustering for pose estimation and Trimmed ICP (TrICP) for refinement. The impact of the individual parts of our method is evaluated in an ablation study on LM.

\textbf{Datasets:} 
The LM dataset \cite{Hinterstoisser2012} is used to evaluate the single-object setting. It consists of 15 scenes with individual toys and household objects. A test set is defined based on the BOP19 challenge \cite{BOP}, albeit adapted to learning-based methods. These methods use the training split defined in \cite{Brachmann2016,Rad2017,Tekin2018}, which excludes scenes 3 and 7 but includes 15\% of the test frames used in \cite{BOP}. We therefore exclude both scenes and the frames used in training from the test set for a total of 2219 test frames.
xAPC \cite{Mitash2018} and YCBV \cite{Xiang2017} are used for the multi-object setting. Both datasets exhibit clutter as well as isolated, 2- and 3-object support relationships. xAPC uses Amazon Picking Challenge objects and features three objects per scene. The whole dataset is used for testing.
YCBV contains 92 scenes. The 12 test scenes consist of 3 to 6 objects from the YCB object set~\cite{Calli2015}. The test set defined in \cite{BOP} is used for our evaluation.

\textbf{Metrics:} The procedure defined for the BOP 2019 challenge \cite{BOP} is used for evaluation. This considers three different error functions, namely, the Maximum Symmetry-Aware Projection Distance (MSPD), the Maximum Symmetry-Aware Surface Distance (MSSD) and the Visible Surface Discrepancy (VSD). 
The reported values per error function are the average recall rates over 10 thresholds in percent. The overall performance score (AR) is the average recall rate over all sub-scores.
On xAPC, we additionally report the average rotation and translation errors for comparison with \cite{Mitash2018}.

\textbf{Baselines:} 
Mitash et al.~\cite{Mitash2018} (PHYSIM-MCTS) evaluate on the xAPC dataset. For comparability, we use the code provided by the authors to generate bounding boxes, a pool of 25 hypotheses per object and the results reported for their method. A maximum of 150 TrICP iterations is used for evaluation of all approaches. Note that, for PHYSIM-MCTS, we only count the refinement iterations in the expansion step to ensure a fair comparison.
The best performing methods on LM are the PPF-based methods by Vidal et al. \cite{Vidal2018} and Drost et al. \cite{Drost2010}. As neither provide code, we use the code of a comparable PPF-based method by Alexandrov et al.~\cite{Alexandrov2019} to produce a pool of hypotheses. We train Mask R-CNN \cite{He2017} to provide detections and segmentation masks.
In addition, we evaluate the RGB-D-based method DenseFusion~\cite{Wang2019}. It features a fast inference time and a learning-based refinement method. Precomputed detections and segmentation masks by~\cite{Xiang2017} are used. A pool of object pose estimates is generated using the provided code and weights. The hypotheses pool consists of the highest confidence per-pixel estimate and additional uniformly-random sampled estimates.

We set the parameters for the verification score in Equation~\eqref{eq:score} to $\tau=20$mm and $\alpha=45$deg on all datasets. PyBullet~\cite{pybullet} is used as physics simulator with a time-step of 1/60sec, 10 solver iterations, 4 sub-steps and assuming an equal mass of 1kg for all objects. 3D plane segmentation is employed to determine a supporting plane and the normal is used to compute the gravity direction.

The generality of our approach is shown by applying it to three baseline iterative refinement approaches, namely, TrICP, point-to-point ICP and DF-R.
TrICP uses the implementation in PCL \cite{pcl} with the same settings as \cite{Mitash2018}. The simulation uses 60 steps in this case.
We use the basic point-to-point ICP implementation from PCL with 50 iterations. For our approaches, we distribute the ICP iterations evenly over 5 PIR iterations.
DF-R uses the weights provided by~\cite{Wang2019}, trained to use 2 iterations. They are distributed over 2 PIR iterations. As ICP and DF-R are more sensitive to interference with the iterative refinement procedure, only 3 simulation steps are used.


\subsection{Ablation Study}

The following ablations aim to motivate several design choices. The experiments start with the ground-truth annotations of the LM dataset as initial estimates and introduce errors of increasing magnitude. For the ablation, the ground-truth ground plane is used for physics simulation. Two types of errors are applied. (1) Rotation error is created by uniformly-random sampling a rotation axis from the unit sphere and rotating the ground-truth estimate by a varying angle about this axis. (2) Translation error is introduced by offsetting the ground truth by a translation vector that is sampled from the unit sphere, scaled by a varying distance.

\begin{figure}[!t]
\centering
\vspace{1ex}
\includegraphics[width=\linewidth]{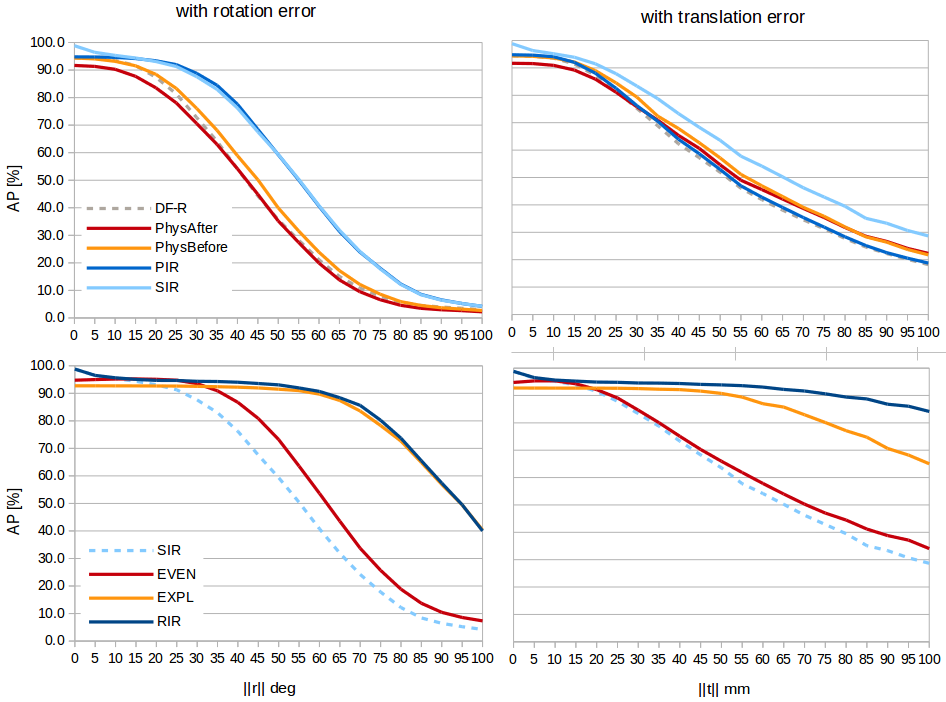}
\caption{Ablations on LM using single hypotheses (top) and 5 hypotheses (bottom). EVEN and EXPL use our verification score to determine the best estimate and PIR for refinement. PhysBefore and PhysAfter apply simulation before and after refinement. AR values at 5mm/deg steps are reported and interpolated in between.}
\label{fig:ablation}
\end{figure}

\subsubsection{Physics Simulation and Iterative Refinement}
As shown in Figure \ref{fig:ablation} (top), our interleaved approach to combine physics simulation with refinement (PIR) is consistently the best performing simulation approach under rotation error. For translation error, it is limited as it only considers the rotation part from simulation to contain divergence.
The benefit of using only rotation is illustrated by comparison with applying full simulation after refinement (PhysAfter) as used in~\cite{Mitash2018}. Rotation error in the initial estimate causes this approach to diverge and perform even worse than the baseline method (DF-R) without physics simulation. Figure \ref{fig:ablation} (top) also shows the benefit of supervising the refinement process. Our approach (SIR) consistently improves the accuracy of pose estimates, most notably under translation error.

\subsubsection{Regret Minimization}
There are two major approaches to deal with multiple hypotheses. The first is to score all initial hypotheses, exploiting only the best scoring hypothesis for refinement (EXPL).
The second approach is to refine all hypotheses evenly and selecting the best scoring refined hypothesis (EVEN).
As shown in Figure \ref{fig:ablation} (bottom), EVEN performs well for low error magnitudes while EXPL is robust to high error magnitudes. Our regret-minimizing approach (RIR) balances between these two extreme approaches and is thus able to outperform the alternatives. Moreover, a comparison with SIR shows the benefit of considering multiple hypotheses.

\subsection{Robustness Analysis}
To highlight the robustness of our approach, we perform experiments with missing depth values to consider two types of errors.
(1) Occlusion is simulated by removing rectangular patches that are centered at uniformly-random sampled positions of the observed object. (2) Missing parts of objects from the depth channel, e.g., due to reflective material, are considered by removing depth values that correspond to the object above a certain height.
Error is introduced similar to the ablation study but kept fixed at 5mm for translation and 5deg for rotation. The depth error increases from 0 to 90\%.

\begin{figure}[!t]
\centering
\vspace{1ex}
\includegraphics[width=\linewidth]{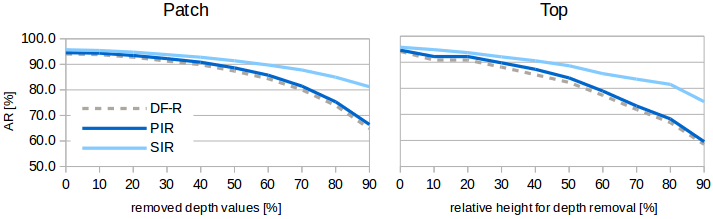}
\caption{Robustness study on LM using single hypotheses with a fixed error magnitude of 5mm and 5deg. AR values are measured every 10\% and interpolated in between.}
\label{fig:robustness}
\end{figure}

As shown in Figure \ref{fig:robustness}, our approach increases robustness to both types of error in comparison to the baseline. This indicates that the remaining depth information, together with physics simulation, limit the degradation of performance.

\subsection{Comparison to State of the Art}
\subsubsection{Single-Object Scenario}


\begin{table}[!t]
    \scriptsize
    \centering
    \caption{Comparison using DF-R\cite{Wang2019} and ICP\cite{Besl1992} on \texttt{LINEMOD}.}
    \begin{tabular}{l||c|c|c|c||c|c}
        DF         &   VSD     & MSPD  & MSSD  & AR    & T[ms] & \#ref/obj \\ \hline\hline
        \cite{Wang2019}     &   70.6    & 76.8  & 77.7  & 75.0  & 2 & 2      \\ \hline
        PIR         &   73.3    & 79.3  & 80.1  & 77.6  & 4 & 2      \\ \hline
        SIR         &   74.0    & 85.9  & 86.4  & 82.1  & 14 & 2      \\ \hline\hline
        \cite{Wang2019}     & 76.9      & 82.6  & 82.9  & 80.8  & 11 & 10      \\ \hline
        RIR         &   \textbf{78.3}    & \textbf{89.7}  & \textbf{89.6}  & \textbf{85.9}  & 48 & 10      \\ \\
        PPF         &   VSD     & MSPD  & MSSD  & AR    & T[ms] & \#ref/obj \\ \hline\hline
        \cite{Besl1992}     &   79.8    & 93.2  & 93.0  & 88.7  & 248 & 50      \\ \hline
        PIR         &   78.1    & 92.1  & 92.1  & 87.4  & 274 & 50      \\ \hline
        SIR         &   79.9    & 93.7  & 93.2  &  88.9 & 302 & 50      \\ \hline\hline
        \cite{Besl1992}     &   80.0    & 93.4  & 93.2  & 88.9  & 617 & 150     \\ \hline
        RIR         &   \textbf{81.0}    & \textbf{95.1}  & \textbf{94.5}  & \textbf{90.2}  & 892 & 150     \\ 
    \end{tabular}
    \label{tab:results_lm}
\end{table}

The single-object scenario is evaluated on the LM dataset using DF and PPF as object pose estimators and DF-R and ICP as refinement methods. The refiners are run for the same number of iterations for comparison with RIR. Results are shown in Table~\ref{tab:results_lm}.

The performance of PIR indicates that physics simulation is beneficial given less accurate initial estimates using DF as compared to PPF. This agrees with our hypothesis that simulation improves implausible initial estimates while being vulnerable to divergence in inaccurate simulation environments.
In both conditions, the biggest relative improvement is achieved by SIR, improving over DF-R by 7.1\% AR. As indicated by the results using PPF, SIR is able to limit the divergence of the physics simulation observed for PIR.
The top-performing approach in both conditions is RIR, improving over the baselines using the same number of refinement iterations by 5.1\% and 1.3\% AR, respectively.

Regarding runtime, we observe the application of physics simulation results in a small relative increase per frame of 1ms per simulation. Note that, using DF-R as refinement method, SIR and RIR still achieve 71fps and 21fps.


\subsubsection{Multi-Object Scenario}

An evaluation on the YCBV and xAPC datasets highlights the performance in multi-object scenarios. For comparison with RIR, VF$_b$ and VF$_d$, the baseline DF-R is also run for the same number of iterations.

The results on YCBV are shown in Table~\ref{tab:results_ycbv}.
The supervision through SIR is again the biggest source of relative improvement as compared to DF-R with an increase of 2.6\% AR.
The increased number of refinement iterations decreases the performance of DF-R. This could be due to the confidence score of DF suggesting a sub-optimal initial estimate for exploitation or due to divergence of the refinement method itself. In either case, RIR does not exhibit divergent behavior and is able to outperform the baseline method given the same number of refinement iterations by 6.5\%. Table \ref{tab:results_exapc} shows that on xAPC, the performance of RIR improves over the approach by Mitash et al.~\cite{Mitash2018} on the VSD and MSPD metrics by 3.3\% and 0.4\% and significantly speeds-up the runtime.

The results on both datasets show that our scene-level approaches successfully deal with the occlusion and support relationships in multi-object scenarios. Both VF$_b$ and VF$_d$ outperform~\cite{Mitash2018} by a significant margin of 2.3\% and 5.0\% AR, respectively, with VF$_d$ performing the best overall. All our approaches are approximately five times faster, with TrICP accounting for 5s per frame. This highlights the benefit of the initialization of the solutions, the efficient search space formulation and our GPU-based computation of the verification score.
As YCBV contains highly cluttered scenes that introduce occlusion and features few support relationships, the relative increases over RIR are less pronounced with 0.1\% and 0.3\%. Overall, our scene-level approaches perform best on YCBV with VF$_b$ achieving an increase of 6.8\% over DF-R given the same number of iterations.

\begin{table}[!t]
    \scriptsize
    \centering
    \vspace{1ex}
    \caption{Comparison with Mitash et al. \cite{Mitash2018} using Trimmed ICP \cite{pcl} on \texttt{Extended APC} with 150 iterations each.}
    \begin{tabular}{l||c|c|c|c||c|c||c}
        PCS         &   VSD     & MSPD  & MSSD  & AR    & $\Bar{r}$ [deg]  & $\Bar{t}$ [cm]   & T[s] \\
        \hline
        \hline
        \cite{Mitash2018}        &   48.5    & 51.6  & 68.3  & 56.2  & \textbf{5.7}  & 1.3   & 29.9     \\ \hline
        RIR         &   51.8    & 52.0  & 63.0  & 55.6  & 10.5  & 1.4   & 5.5      \\ \hline
        VF$_b$      &   54.4    & 54.3  & 66.7  & 58.5  &  8.0  & \textbf{1.2}   & 5.5      \\ \hline
        VF$_d$      &   \textbf{56.7}    & \textbf{57.3}  & \textbf{69.6}  & \textbf{61.2}  &  7.5  & \textbf{1.2}   & 6.2      \\ 
    \end{tabular}
    \label{tab:results_exapc}
\end{table}

\begin{table}[!t]
    \scriptsize
    \centering
    \caption{Comparison using DF-R\cite{Wang2019} on \texttt{YCB-VIDEO}.}
    \begin{tabular}{l||c|c|c|c||c|c}
        DF          &   VSD     & MSPD  & MSSD  & AR    & T[ms] & \#ref/obj \\
        \hline
        \hline
        \cite{Wang2019}     &   74.2    & 69.9  & 77.6  & 73.9  & 17 & 2      \\ \hline
        PIR         &   74.9    & 70.8  & 78.2  & 74.7  & 20 & 2     \\ \hline
        SIR         &   76.5    & 72.9  & 80.2  & 76.5  & 49 & 2     \\ \hline\hline
        \cite{Wang2019}     &   71.2    & 66.3  & 75.6  & 71.0  & 71 & 10      \\ \hline
        RIR         &   77.9    & 73.9  & 80.6  & 77.5  & 228 & 10      \\ \hline
        VF$_b$      &   78.3    & 73.8 & 80.6  & 77.6  & 495 & 10      \\ \hline
        VF$_d$      &   \textbf{78.5}    & \textbf{74.1}  & \textbf{80.9}  & \textbf{77.8}  & 521 & 10      \\
    \end{tabular}
    \label{tab:results_ycbv}
\end{table}


\subsection{Robotic Grasping Experiment}
\begin{figure}[!t]
\vspace{1.0ex}
\centering
\includegraphics[trim=285 170 115 100, clip,width=0.3\linewidth]{figures/teaser_render.png}
\includegraphics[trim=40 340 20 212, clip,width=0.3\linewidth]{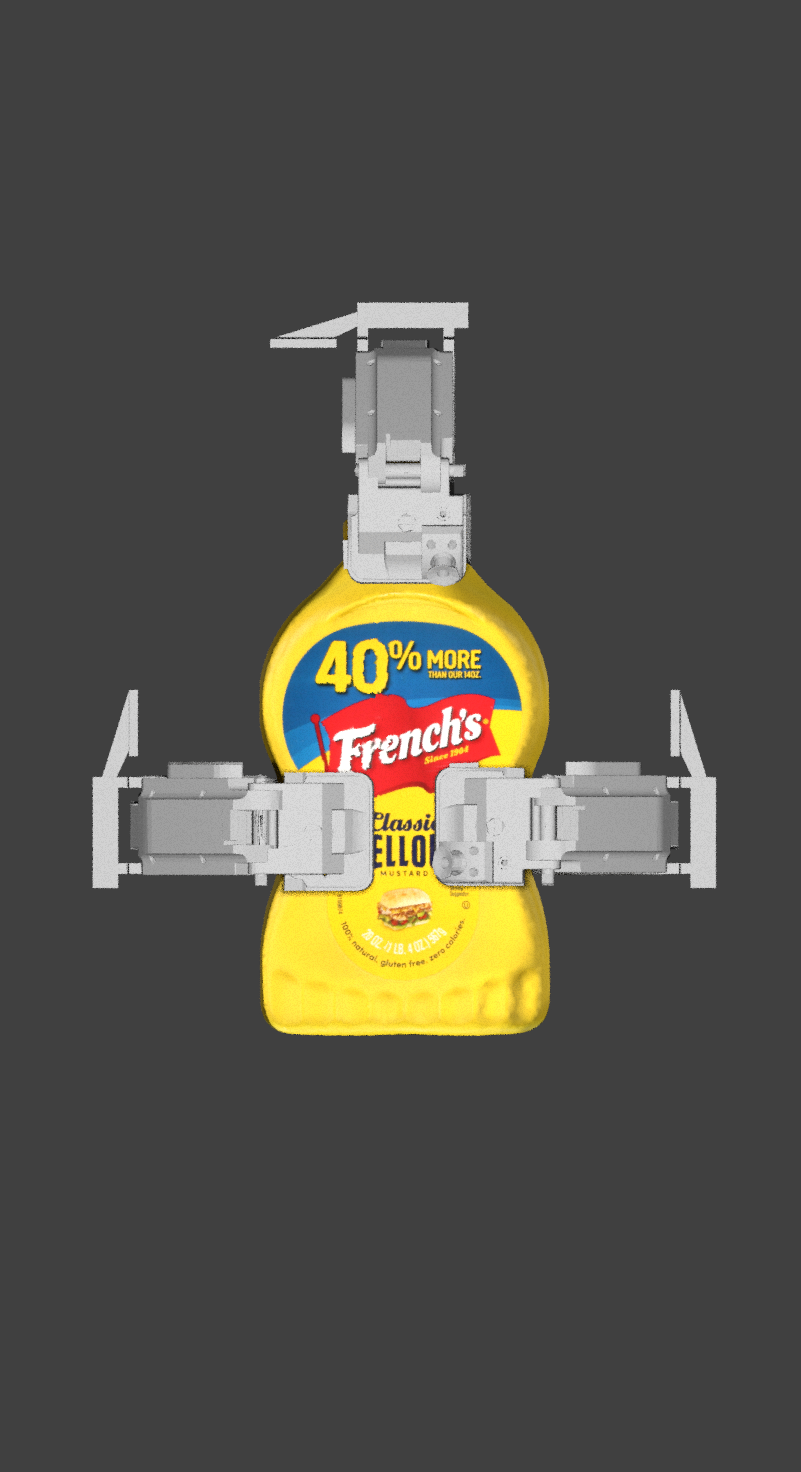}
\includegraphics[trim=0 7 0 0, clip,width=0.3\linewidth]{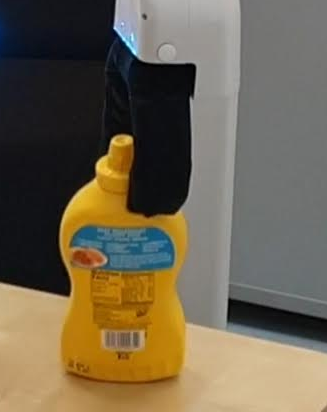}
\caption{Refined estimates using RIR (left), retrieved annotated grasps (mid) and successful grasp attempt (right).}
\label{fig:grasping}
\end{figure}

Our work is motivated by the performance deterioration of object detection and pose estimation methods when deployed on robots \cite{Loghmani2018,Ammirato2017}. To evaluate whether the proposed approach is able to reduce this problem, its performance is evaluated in a grasping experiment using a Toyota HSR and YCBV objects. Reproducible experimental conditions are ensured by using the GRASPA scene layouts~\cite{Bottarel2020} to place 5 objects as shown in Figure \ref{fig:teaser}. 10 grasps are attempted per object -- 5 are attempted for a given pose and an additional 5 for a rotation to a symmetric pose. Multiple grasp poses are annotated by hand for each object as shown in Figure~\ref{fig:grasping} (mid).

In each experiment, Mask R-CNN~\cite{He2017} is executed to detect objects and to provide instance segmentation masks. The evaluated methods are queried to compute an object pose estimate from this information and the RGB-D image. Using this pose estimate, the annotated grasp poses are transformed to the scene, then checked for collision with the octomap. Trajectories for all collision-free grasps are planned using MoveIt \cite{Sucan2013}. If at least one plan is found, this is counted as a \textit{found} grasp. A grasp is considered a \textit{success}ful grasp if the plan can be executed, i.e., the object is grasped and remains stable in the robot's gripper.

As shown in Table \ref{tab:results_grasping}, our proposed approach generates object pose estimates that result in more successful and reliable grasps. The most striking improvements are achieved on the ``061\_\textit{foam}\_brick'' and ``011\_\textit{banana}'' objects. Due to their proximity to other objects, object poses must be accurate to allow collision-free grasps. The \textit{banana} is the most difficult object, resulting from inaccuracy in the instance segmentation and the low height.

\begin{table}[!t]
    \scriptsize
    \centering
    \vspace{1ex}
    \caption{Results of grasping experiments in percentage of \textit{found} collision-free grasp poses and \textit{success}ful grasps.}
    \setlength{\tabcolsep}{4pt} 
    \begin{tabular}{l||c|c|c|c|c||c|c|c}
        DF                  & mustard & spam & foam & jello & banana & success & found & \#ref/obj \\
        \hline
        \hline
        \cite{Wang2019}     & 10 &  3 &  1 &  7 &  0 & 42\% & 46\% &  2 \\ \hline
        SIR                 &  9 &  7 &  2 &  7 &  0 & 50\% & 70\% &  2 \\ \hline\hline
        \cite{Wang2019}     & 10 &  6 &  5 &  9 &  1 & 62\% & 70\% & 10 \\ \hline
        \cite{Mitash2018}   &  9 & 10 &  2 &  6 &  0 & 54\% & 78\% & 10 \\ \hline
        RIR                 & \textbf{10} & \textbf{10} &  \textbf{9} & \textbf{10} &  \textbf{4} & \textbf{86\%} & \textbf{90\%} & 10 \\ \hline
    \end{tabular}
    
    \label{tab:results_grasping}
\end{table}

\section{Conclusion}\label{sec:conclusion}
This work presented an approach for the tight integration of hypotheses verification, refinement and physics simulation for object pose estimation. The rendering-based hypotheses verification and the proposed physics-guided extension to iterative refinement methods benefit from this integration by allowing them to share useful information. The comparison with state-of-the-art methods and a robotic grasping experiment show that our integrated approach creates more accurate and more reliable object pose estimates. Furthermore, we are able to increase performance over related work while significantly reducing the runtime.

An open issue for robot systems is the presence of a-priori unknown objects. With interactions between known and unknown objects, the results of simulation will diverge from the true object pose. In these cases, incorporating shape estimation would enable unknown objects to be considered in simulation.
Moreover, the use of physics simulation requires fixed structures on which objects can rest and an estimate for the gravity vector. For robotic applications, static objects with non-planar surfaces in the robot's environment map could be considered as supporting structures. An IMU could be used to determine the gravity direction to become robust to tilted or non-planar support.

% ...


\end{document}